\documentclass[conference]{IEEEtran}
\ifCLASSINFOpdf
   \usepackage[pdftex]{graphicx}
   \graphicspath{{../pdf/}{../png/}}
   \DeclareGraphicsExtensions{.pdf,.jpeg,.png}
\else
\fi

\usepackage{siunitx} 

\usepackage[caption=false,font=footnotesize]{subfig}
\usepackage{fixltx2e}
\begin{document}
%
\title{Model-Agnostic Defense for Lane Detection against Adversarial Attack}


\author{\IEEEauthorblockN{Henry Xu}
\IEEEauthorblockA{University of California, Berkeley\\
henryxu@berkeley.edu}
\and
\IEEEauthorblockN{An Ju}
\IEEEauthorblockA{University of California, Berkeley\\
an\_ju@berkeley.edu}
\and
\IEEEauthorblockN{David Wagner}
\IEEEauthorblockA{University of California, Berkeley\\
daw@cs.berkeley.edu}}


%



\IEEEoverridecommandlockouts
\makeatletter\def\@IEEEpubidpullup{6.5\baselineskip}\makeatother
\IEEEpubid{\parbox{\columnwidth}{
    Workshop on Automotive and Autonomous Vehicle Security (AutoSec) 2021 \\
    21 February 2021 \\
    ISBN 1-891562-68-1 \\
    https://dx.doi.org/10.14722/autosec.2021.23032 \\
    www.ndss-symposium.org
}
\hspace{\columnsep}\makebox[\columnwidth]{}}

\maketitle



\begin{abstract}

Susceptibility of neural networks to adversarial attack prompts serious safety concerns for lane detection efforts, a domain where such models have been widely applied. Recent work on adversarial road patches have successfully induced perception of lane lines with arbitrary form, presenting an avenue for rogue control of vehicle behavior. In this paper, we propose a modular lane verification system that can catch such threats before the autonomous driving system is misled while remaining agnostic to the particular lane detection model. 
Our experiments show that implementing the system with a simple convolutional neural network (CNN) can defend against a wide gamut of attacks on lane detection models.
With a 10\% impact to inference time, we can detect 96\% of bounded non-adaptive attacks, 90\% of bounded adaptive attacks, and 98\% of patch attacks while preserving accurate identification at least 95\% of true lanes, indicating that our proposed verification system is effective at mitigating lane detection security risks with minimal overhead.
\end{abstract}


%

\section{Introduction}

\begin{figure}[!t]
\centering
\includegraphics[width=3.5in]{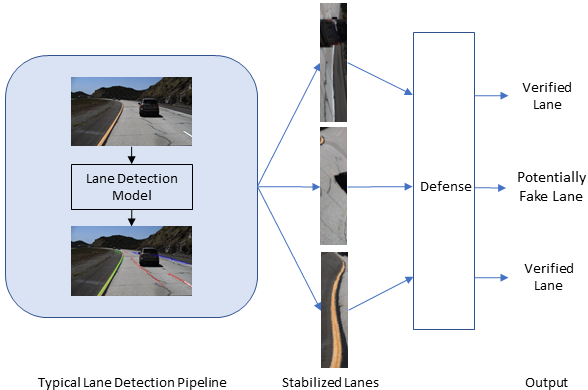}
\caption{Our proposed defense augmentation to a general lane detection model.}
\label{fig_flow}
\end{figure}

End-to-end lane detection methods have shown great promise; however, their shared foundation with deep neural networks imply a shared weakness to adversarial examples~\cite{adversarialExampes}. Given the importance of accurate lane detection in downstream control decisions for autonomous vehicles, a successful attack on lane perception could result in undesirable or outright dangerous vehicle behavior.
In particular, we are interested in attacks that could interfere with vehicle guidance through the generation of malicious lane lines, where attack success is marked not by alarm, as is the case when lane lines cannot be found, but by a false sense of normalcy. With no defense, as is the case with current state-of-the-art efforts, a lane detection pipeline is unable to make any judgement of lane validity, and thus the perceived fake lanes are indistinguishable from real. 
To defend against such attacks, we propose a system for lane verification as illustrated in Figure \ref{fig_flow}, with the goal not to recover the original lanes, but to minimize instances of lane detection model false confidence.

Our lane verification model is fast, lightweight, and applicable to any existing lane detection effort.
The simplicity of our verification model imparts very little inference overhead, and its modular nature allows for independent training that avoids the costs associated with redesigning and retraining the large and complex neural networks commonly seen in industrial lane detection systems.   

The modularity carries the additional benefit of being lane detection model-agnostic, paving a path for integration into any lane detection pipeline.
Given the constant improvement and refinement of lane detection techniques, detaching our defense from a particular architecture allows it to remain viable as the underlying methods become more sophisticated.

Our system is motivated by the framing of secure lane detection as two complementary tasks: lane detection and lane verification. The former requires discerning the locations of a variable number of lanes in a constantly changing environment; the latter boils down to binary classification: given a set of lane coordinates, determine if they correspond to a lane that is either real or fake. 
Instead of further complicating the optimization problem faced by existing lane detection models by introducing the secondary goal of security on top of their initial purpose, we propose moving the task of verification into a separate bespoke model, allowing each part of the pipeline to focus on maximizing individual performance without compromise. 
Since the task of lane verification can take lane coordinates as given and only needs to return a binary result, it can be accomplished by models much simpler and faster than those required for lane detection.

\begin{figure}[!t]
\centering
\includegraphics[width=3.5in]{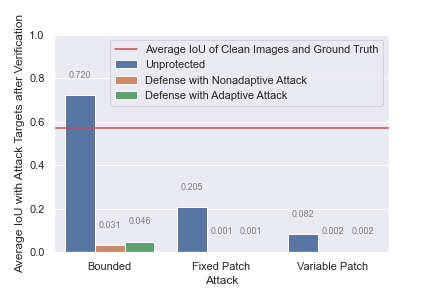}
\caption{Defense Performance. Our defense is able to filter out all but a handful of adversarial attacks that would otherwise fool unprotected lane detection models. Intersection over Union (IoU) between detected lanes and attack targets is our chosen attack metric due to its measurement of both how well the induced fake lane matches the target fake lane and how much of the original scene was preserved, providing a sense of amount of control an attacker has over the scene. The bounded attack before verification surpasses the IoU achieved by the lane detection model on real lanes, suggesting the adversarial fake lanes are indistinguishable from real ones to the model. While patch attack IoU is relatively muted compared to that of bounded attack due to its largely local impact, it can still cause significant lane deviation as shown in Figure \ref{fig_patch_attack}.  }
\label{fig_iou_chart}
\end{figure}

Our experiments show that simple convolutional models are sufficient to significantly improve lane detection pipeline robustness to both digital and physical attack as pictured in Figure \ref{fig_iou_chart}. When evaluated against a $L_p$ bounded attack and two patch-based attacks, our model can detect over 95\% of attacks with minimal impact to model accuracy and inference time. These results suggest that our model is capable of defending against a variety of attack types, including unknown threats.


In summary, strong performance of our defense against both nonadaptive and adaptive versions of such threats indicates that such a system could offer security to lane detection models at very little expense. 

Our primary contributions are as follows:
\begin{itemize}
    \item We propose a simple lane verification defense that can be integrated into the pipeline of any lane detection effort with no retraining of the underlying model required. Its independent and lightweight nature provides marginal inference overhead and allows for quick security updates when new attacks arise. 
    \item We show empirically that verification provides significant lane detection security with minimal cost. 
\end{itemize}

\section{Related Work}
While work on lane detection model defenses is sparse, there is extensive related work on end-to-end detection models and some work on lane detection attacks which have been summarized below.

\subsection{End-to-End Lane Detection Models}
Convolutional neural network-based lane detection models typically frame the core task as one of image segmentation, with the goal to label each pixel as one of $N$ classes, each class corresponding to a distinct lane. In the end-to-end formulation, the segmentation is accompanied by a parallel binary labelling of lane existence, allowing the model to narrow down where exactly the lane is within pixels of the same segmentation class. By doing so, an end-to-end lane detection model is able to take a scene and return predictions of lane line locations. Proposed models largely differ on neural network architecture choices and postprocessing cleanup procedures, the details of which our defense treats as a black box. 

Given the methodological similarities between the top lane detection models on the TuSimple dataset, we chose to test our proposed defense with LaneNet as proposed by Neven et al. \cite{LaneNet} due to its near state-of-the-art performance at time of writing and result reproduction accessibility. We achieve accuracy within $2\%$ of Neven et al.'s results before adding our defense. Note that due to the lane detection model-agnostic nature of our defense, the results from our experiments should be applicable to any other model we could have chosen, such as \cite{hou2019learning}, \cite{jung2020lightweight}, and \cite{ko2020key}. 

\subsection{Image Comprehension Model Attacks and Defenses}
The framing of lane detection as a task of image segmentation suggests a sharing of similar security weaknesses, and recent work has shown image comprehension models to be very susceptible to adversarial attack. Adversarial examples, shown to be incredibly effective for image classification, have been shown to be extendable to image segmentation, with \cite{cisse2017houdini} specifying a framework generalizing their generation across a variety of tasks, including segmentation, and \cite{adversarialExampes} finding attack success across a variety of segmentation networks. Successful attack need not change every pixel as discovered in \cite{brown2018adversarial}, and classification can easily be corrupted with a patch a fraction of the total size of the image. Defenses, such as adversarial training, as suggested and explored by \cite{goodfellow2015explaining}, \cite{nokland2016improving}, \cite{madry2018towards}, and \cite{pmlr-v97-wang19i} against adversarial attack, often involve retraining the entire model  which is costly given the ever increasing complexity of state-of-the-art techniques.

\subsection{Lane Detection Model Attacks}
Regarding adversarial attacks on lane detection models in particular, recent work has used image segmentation attack methods to great effect. \cite{sato2020hold} details how a bounded patch, disguisable in practice as road dirt, could be used to fool lane detection models before the passenger catches on. The adversarial patches we test our defense on differ in their much smaller size and unboundedness compared to the full lane covering required by \cite{sato2020hold}. Although our metric of attack success show the patches are unable to achieve our goal of reshaping all the lanes in the scene, our results do reaffirm that patch attack-based lane deviation is a valid threat necessitating defense.

\begin{figure}[!t]
\centering
\includegraphics[width=3.5in]{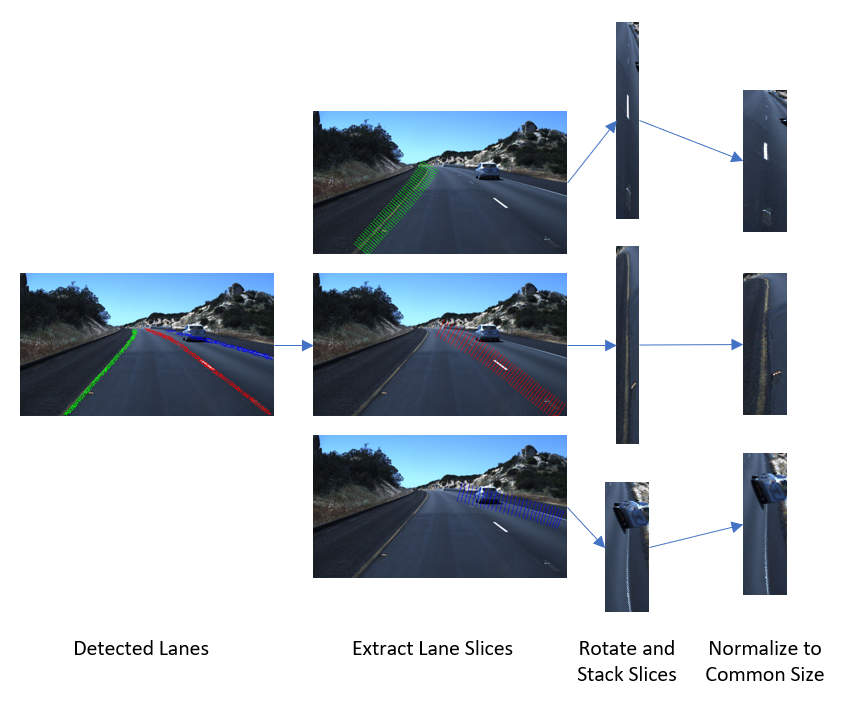}
\caption{Lane Stabilization. Stabilized lanes are classified as either real or fake by the verification system.}
\label{fig_stabilization}
\end{figure}

\section{Method}

\subsection{Defense}
Our proposed defense takes place at the end of an existing lane detection pipeline, at which point a set of candidate lanes have been identified by the lane detection model. Upon attack success, these candidate lanes are corrupted and may include a mix of real and fake lanes. The goal of our defense is to verify the real lanes and filter out the fake lanes before further autonomous vehicle systems make potentially hazardous decisions based on faulty information.  


\subsubsection{Stabilizing Lanes}
\begin{figure}[!t]
\centering
\includegraphics[width=3in]{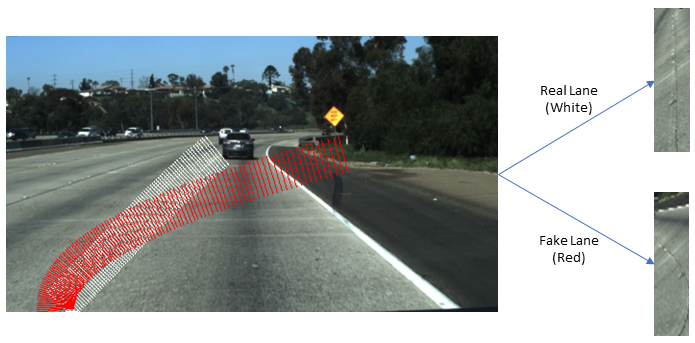}
\caption{Examples of positive and negative examples generated by the lane stabilization process for the defense training dataset.}
\label{fig_defense_dataset}
\end{figure}

The defense takes detected lanes as input; however, due to the nature of perspective and exacerbated by the fact that lane lines can curve either left or right, extracting lanes using masks formed from the pixel-level segmentation as provided by the lane detection model results in lane lines that can take an arbitrary number of forms, lending itself to a classification problem with an unbounded domain. To address this issue, we construct a stabilized image of each lane as follows: 
\begin{enumerate}
    \item Given a set of points corresponding to a lane, we first perform a least squares polynomial regression to get its underlying shape. 
    \item For each pixel that lies on the curve, we compute the derivative of the polynomial at that point, and extract the pixels corresponding to the line centered around the point on the curve and rotated by the angle formed by the derivative and the horizontal axis. 
    \item Rotating each extracted line such that the pixels in each line are lined up horizontally, we can then stack the horizontal lines vertically to obtain a stabilized image of the lane that is not influenced by perspective or lane curvature. 
\end{enumerate}
While inverse perspective transforms~\cite{massimo1998stereo} can mitigate perspective distortions as well, we found little benefit from applying a fixed homography before extracting the stabilized lanes. Figure \ref{fig_defense_dataset} shows examples of positive and negative samples generated by our stabilization process.

\subsubsection{Training}
Given the absence of applicable existing datasets, to generate a training dataset for the defense, we extract stabilized real and fake lanes from a dataset labelled for lane detection, using the ground truth labels as the basis for real lanes and generating curves that start from real lanes but deviate as they move away from the camera as the basis for fake lanes. While each scene only has a fixed number of real lanes, an arbitrary number of fake lanes can be generated from them, creating an imbalance in real versus fake lanes. A potential enhancement of the defense could involve tailoring fake lane design in anticipation of specific attacks.

To address class imbalance, the defense is trained using focal loss as described in \cite{FocalLoss} with hyperparameters determined experimentally. To improve robustness, we employ adversarial training~\cite{madry2018towards} with an emphasis on missed fake lane detection via asymmetric weighting of the loss function.



\subsection{Nonadaptive Threat Model}
We propose two different threat models dependent on attacker access to lane detection equipment. The threats share the goal of perturbing the original scene such that the binary segmentation component of an end-to-end lane detection model is disrupted, allowing the attacker to inducing an arbitrary lane existence pattern of their choosing. Due to the downstream dependence of lane detection models on lane existence, the attack can fatally disrupt lane detection model function. All attacks are carried out using Projected Gradient Descent~\cite{madry2018towards} until convergence.

The attacks are nonadaptive, meaning that they do not target our verification model. We describe adaptive versions of our attacks~\cite{tramer2020adaptive} in the following section.
\subsubsection{L-Infinity Attack}
In our first threat model, the attacker is able to manipulate any pixel in the input image, but is constrained by a bound on the size of perturbation for each pixel, akin to digital corruption of the input or a lens filter over the camera. 
\subsubsection{Patch Attack}
The attacker is able to manipulate a subset of pixels in the input image, but is not constrained by a bound on the size of perturbation for each pixel. We can further break the subset of pixels into two sizes: fixed and variable. In a fixed size patch attack, there is a fixed number of pixels the attacker can modify, akin to having a patch on the lens of the camera. In a variable size patch attack, the number of pixels able to be modified is a function of distance away from the camera, akin to being able to lay a physical patch on the road in the scene. We can simulate such an attack by scaling patch size against how lane width and lane marker height change as a function of pixel height.

\subsection{Adaptive Threat Model}
The lane stabilization procedure poses an issue for differentiability when planning an adaptive attack in which the attacker seeks to outmaneuver the defense. Specifically, the backpropagation when attacking the defense is unresolvable at the point of lane stabilization. We instead propose an adaptive threat model that takes place in two stages that can extend any of the nonadaptive threat models proposed above. Without loss of generality, we discuss the adaptive threat in the context of an L-infinity attack. 
\begin{enumerate}
    \item The first stage is an L-infinity attack on the lane detection model, with the goal to induce an arbitrary binary segmentation map of our choosing. The output is a perturbed scene in which the lane detection model identifies the arbitrary lane we choose. 
    \item Once the first stage has converged, the pixels corresponding to the arbitrary lane are extracted from the perturbed scene. The lane is then stabilized as described in the process above, and subject to the second stage of the attack. The goal of the second stage is to find a perturbation to the stabilized lane such that the defense is fooled into thinking the arbitrary lane is real. Upon convergence of the L-infinity attack on the defense, the resulting perturbation on the stabilized lane is mapped back to the original location of the pixels that form the stabilized lane in the perturbed scene output from the first stage. 

\end{enumerate}
    The final result is a perturbed scene designed to both convince the lane detection model of the existence of a fake lane and the defense that the perceived fake line is real.

\section{Experiments}
\subsection{Datasets}
The lane detection model, LaneNet, is trained on the TuSimple dataset \cite{tusimple}, which consists of 3,626 training images and 2,782 testing images taken from a camera mounted on the front of a vehicle. The images are scaled to be of size 512x288. 

The defense is trained on stabilized lanes extracted from the TuSimple dataset. Each stabilized lane is fit with a polynomial of degree 3 and resized to be of size 128x40. 
\begin{figure}[!t]
\centering
\includegraphics[width=3.5in]{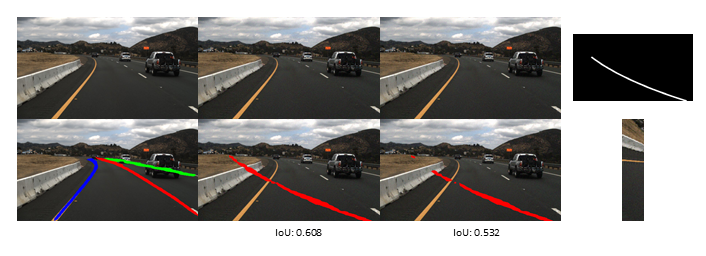}
\caption{Example of L-infinity bounded attack. From left to right, the vertical pairs of images represent clean, nonadaptive, and adaptive attack, with the lane detection model output superimposed on the second row, and IoU values below. IoU values are with respect to the upper rightmost image of the target lane binary segmentation map, and the resulting stabilized lane is shown below. Under bounded attack, the attacker is able to assert full control over the scene and induce arbitrary lane configurations.}
\label{fig_bounded_attack}
\end{figure}

\begin{figure}[!t]
\centering
\includegraphics[width=3in]{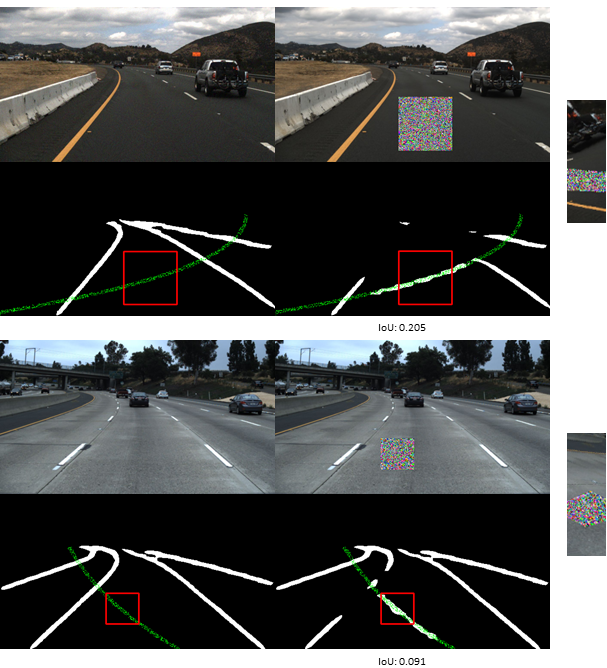}
\caption{Examples of fixed and variable size adaptive patch attacks. Each group of four images has the original and applied patch on the first row, and the corresponding binary segmentation maps on the second. The target lane is dotted in green, with the patch's bounding box outlined in red. IoU values are below each applied patch segmentation map, with the resulting stabilized lane shown to the right of each group. While full scene control is limited, the results are substantial enough to cause rogue vehicle behavior.}
\label{fig_patch_attack}
\end{figure}

\subsection{Implementation Details}




The defense model takes the form of a 3-layer convolutional neural network, intentionally designed to contrast against the numerous layers and weights used for lane detection segmentation. Inputs to the defense take the form of stabilized lanes of size 128x40. The network is comprised of two convolutional layers and one linear layer. Both convolutional layers use a 3x3 filter size with stride 3 and no padding, with BatchNorm and ReLU applied after each.

Both LaneNet and defense model are trained using one GPU (GTX 1080). LaneNet is implemented as specified in~\cite{LaneNet}, with no modifications for robustness, keeping in line with the modular nature of our defense.

All attacks are performed until convergence is achieved. Parameter details are as follows:
\begin{enumerate}
    \item The L-infinity attack is bounded by a per-pixel perturbation of at most 8/255, where the input image's pixels have a range of $[0,1]$.
    \item The fixed size patch attack is a 100x100 square with no bound on pixel deviations inside the square. The square is centered around a point on the targeted arbitrary lane. 
    \item The variable size patch attack is specified by a 100x100 square at the foot of the camera, corresponding to roughly a 3-foot by 3-foot physical patch. For each scene, an arbitrary distance from the camera is selected, and the square is scaled down accordingly.
\end{enumerate}

\subsection{Evaluation}
We evaluate performance of the defense on the test set of TuSimple. For each scene in the original dataset, LaneNet is used to identify lanes in clean and attacked variants. The identified lanes are stabilized and fed to the defense for classification. We report fake lane missed detection rates (false negative) and real lane misclassification rates (false positive) at a classification threshold such that we see a 5\% real lane misclassification rate in a validation set. Additionally, we report the average intersection over Union (IoU) values between all lanes in attacked scenes and attack targets once flagged lanes have been removed. IoU values were of interest due to their dual purpose of measuring how much of the target was achieved and how much of the original scene was preserved, providing a sense of attacker control over the scene.

\section{Results}

\begin{table}[t!]
  \begin{center}

        \caption{Defense results. False positive rate (FPR) refers to real lanes the defense mistakenly flagged as fake, whereas false negative rate (FNR) refers to fake lanes the defense believed to be real. As a measure of post-defense attack success, the false negative (FN) average IoU is the average IoU between all attacked scenes and targets once flagged lanes have removed.}
                \label{tab:results}
    \begin{tabular}{lc|c|c}

       &  \multicolumn{2}{c|}{\textit{Defense Metrics}} & \textit{Attack Metric}\\
      \textbf{Bounded Attack} & \textbf{FPR} & \textbf{FNR} & \textbf{FN Average IoU}\\
      \hline
    Unprotected & 0 & 1 & 0.720\\
      Nonadaptive  & 0.040 & 0.039 & 0.031\\
      Adaptive  & 0.043 & 0.098 & 0.046\\
\\
       &  \multicolumn{2}{c|}{\textit{Defense Metrics}} & \textit{Attack Success Metric}\\
      \textbf{Patch Attack} & \textbf{FPR} & \textbf{FNR} & \textbf{FN Average IoU} \\
      \hline
    Fixed Size Unprotected  & 0 & 1 & 0.205\\
      Fixed Size Nonadaptive  & 0.039 & 0.005 & 0.001\\
      Fixed Size Adaptive  & 0.042 & 0.005 & 0.001\\ 
       \hline
      Variable Size Unprotected  & 0 & 1 & 0.082\\
      Variable Size Nonadaptive  & 0.053 & 0.017 & 0.002\\
      Variable Size Adaptive  & 0.045 & 0.023 & 0.002\\ 
    \end{tabular}
  \end{center}
\end{table}

\subsection{L-Infinity Attack}

We observe that success is almost certain under bounded attack, with all traces of the real lanes wiped out and the induced fake lanes matching target fake lanes with an average IOU of 0.720. Note that after typical lane detection model postprocessing cleanup procedures, induced fake lane and the target fake lane have almost identical stabilized forms. Examples of scenes and their corresponding binary segmentation maps before and after targeted attack are in Figure~\ref{fig_bounded_attack}.

Table \ref{tab:results} shows defense results under bounded attack. With an unprotected model, fake lanes slip by undetected 100\% of the time. Under nonadaptive attack, the defense is very capable of detecting fake lanes while very rarely mistaking real lanes for fake. We do see some gains in attack strength under adaptive attack; however, we are still able to detect 90.2\% of fake lanes. Table~\ref{fig_roc} shows selected ROC curve data.

\subsection{Patch Attack}

\subsubsection{Fixed Size}
The fixed patch attack results reveal a strong reliance of LaneNet on spatially local features. Unlike the previous L-infinity attack, which could manipulate the entire scene to achieve its goal, the patch attack is unable to induce change outside of a small region around the patch location. An example of the attack can be found in the first half of Figure~\ref{fig_patch_attack}. To improve attack stability, the polynomial curve used to stabilize the lane for the defense is pulled from the target binary segmentation map instead of being fitted from what the patch attack is able to induce. Note that while the patch attack does not do well against our attack success metric of full scene control, the achieved result is still capable of causing undesired lane deviation.  

Table~\ref{tab:results} shows defense results under fixed patch attack. Similar to the bounded attack, we see strong detection rates for nonadaptive attack with a minor drop when subject to adaptive attack. Table~\ref{fig_roc} shows selected ROC curve data for the fixed size adaptive attack.

\subsubsection{Variable Size}
Variable patch attack success follows a similar trend to that of the fixed size patch attack, with the region of effect largely localized around the patch location. An example of the variable patch attack can be found in the second half of Figure~\ref{fig_patch_attack}. The same strategy to improve attack stability as described in the previous section is employed when preparing the lane for verification by the defense.

Table~\ref{tab:results} shows defense results under variable patch attack. Mirroring both previous attacks, robust nonadaptive attack detection rates are accompanied by slightly weaker results when under adaptive attack. Although variable patch sizes are generally smaller than the fixed patch, the former attack is slightly stronger than the latter with respect to false positive and false negative rates despite the smaller attack IoU on the unprotected model, suggesting the fixed patch being a more noticeable anomaly harder for the adaptive attack to overcome. Table~\ref{fig_roc} shows selected ROC curve data for the adaptive attack.

\subsection{Speed}
The addition of our defense saw a marginal impact on pipeline inference time, experiencing a drop in output from 29.8 to 27 frames per second. Due to differences in system configuration, we were unable to achieve the 50 frames per second as presented in \cite{LaneNet}, but as a rough estimate, the 3ms inference time of our defense as measured locally would translate to a drop from 52.6 to 45.5 frames per second using the pipeline timings provided by \cite{LaneNet}.


\begin{table}[t!]
  \begin{center}

        \caption{Selected defense ROC curve results. False positive rate (FPR) refers to real lanes the defense mistakenly flagged as fake, whereas false negative rate (FNR) refers to fake lanes the defense believed to be real. \\$\downarrow$ - lower is better $\uparrow$ - higher is better.}
                \label{fig_roc}

                    \begin{tabular}{llc|c|c|c}
                    && \multicolumn{2}{c|}{\textit{Bounded Attack}} & \multicolumn{2}{c}{\textit{Adaptive Patch Attack}}\\
                    && Nonadaptive & Adaptive & Fixed Size & Variable Size\\
                    \hline
                    FNR at 0.01 FPR  &$\downarrow$ & 0.160 & 0.496 & 0.071 & 0.099\\
                    FNR at 0.02 FPR &	$\downarrow$& 0.075 & 0.210 & 0.021 & 0.056\\
                    FNR at 0.05 FPR	&$\downarrow$ & 0.036 & 0.082 & 0.005 & 0.020\\
                    FNR at 0.10 FPR &	$\downarrow$& 0.026 & 0.048 & 0.003 & 0.010\\
                    Area Under Curve &	$\uparrow$ & 0.984 & 0.971 & 0.996 & 0.993\\
                    
                    \end{tabular}


                        
  \end{center}
\end{table}

\subsection{Ablation Studies}
\subsubsection{Adaptive Attack Cycling}
Given that the second stage of the adaptive attack is unable to assess its impact on the first stage, it is possible that there exists adverse feedback between the two stages. To evaluate the extent of this feedback, we test a bounded attack where the stages are cycled up to four times. Results are in the first half of Table~\ref{tab:ablation}. Comparing them with the figures in Table~\ref{tab:results}, we note that the false positive and false negative rates are slightly better for the defense in the cycled attack, suggesting that while repeated cycling may be helping first stage output, the effect is outweighed by an adverse impact to the efficacy of the second stage. Since polynomial fitting cleans up much of the noise in the binary segmentation map, first stage gains may provide marginal benefits to overall attack strength.  
\subsubsection{Simpler Defense Architecture}
We found that although the philosophy of our defense encourages simple defense designs, purely linear models struggle with the task of lane classification as shown in the second half of Table~\ref{tab:ablation}. This result hints again at the highly local nature of lane detection as previously seen in Figure~\ref{fig_patch_attack}, a property linear layers are less adept at taking advantage of compared to the convolutional layers used in our 3-layer model.

\begin{table}[t!]
  \begin{center}
    \caption{Ablation results. False positive rate (FPR) refers to real lanes the defense mistakenly flagged as fake, whereas false negative rate (FNR) refers to fake lanes the defense believed to be real. As a measure of post-defense attack success, the false negative (FN) average IoU is the average IoU between all attacked scenes and targets once flagged lanes have removed.}
    \label{tab:ablation}
    \begin{tabular}{lc|c|c}
           &  \multicolumn{2}{c|}{\textit{Defense Metrics}} & \textit{Attack Success Metric}\\
      \textbf{Cycled Adaptive Attack} & \textbf{FPR} & \textbf{FNR} & \textbf{FN Average IoU}\\
      \hline
      Bounded & 0.042 & 0.044 & 0.030\\ 
      
      \\
             &  \multicolumn{2}{c|}{\textit{Defense Metrics}} & \textit{Attack Success Metric}\\
            \textbf{Linear Defense} & \textbf{FPR} & \textbf{FNR} & \textbf{FN Average IoU} \\
            \hline
                Nonadaptive Bounded & 0.049 & 0.966 & 0.720\\
                Nonadaptive Fixed Size Patch & 0.045 & 0.879 & 0.180\\
            Nonadaptive Variable Size Patch & 0.052 & 0.921 & 0.075\\

    \end{tabular}

  \end{center}
\end{table}

\section{Conclusion}
Our proposed defense sits on top of existing lane detection models and can defend against adversarial attacks with minimal impact to lane detection capabilities. 

The orthogonal nature of the defense allows it to be trained independently from the underlying lane detection model, eliminating the cost of retraining. Amid the rising complexity of image processing models, our defense can provide security with very little overhead. Taking only lane locations as inputs, the defense does not depend on a particular lane detection model's features or assumptions, streamlining integration into any lane detection pipeline. The lightweight nature of the defense promotes fast inference and quick updates as new attacks arise. 

Under a bounded threat model that is able to fully take over the scene, we show that a simple 3-layer model employing our defense structure on top of LaneNet can detect over 90\% of attacks while maintaining a maximum 5\% impact to clean accuracy. Under a patch-based threat model where attacker control is limited but still capable of causing undesired lane deviation, our defense is able to identify 98\% of abnormal activity while preserving the same 5\% threshold. In such situations where the lane detection model would have otherwise passed fake lanes off as real to the autonomous control system, our model is able to call attention to potentially malicious actors. 

Future work could involve designing a differentiable adaptive attack and collecting performance of our proposed defense on a larger sample of lane detection models. An application of transfer learning could be explored by training the defense on a separate dataset from the lane detection model and examining performance. While our defense is able to alert the vehicle to the presence of an attack, it does not provide guidance on the safest response, which we leave as an open question. Variants in the architecture of the verification system may also be worth investigating; for example, defense model enhanced with domain-specific knowledge of the highly local nature of lane detection could see further improvement in verification accuracy. Finally, a similar system could prove useful for bolstering robustness against corruption.


\section*{Acknowledgment}
This work was supported by generous gifts from the Hewlett
Foundation through the Center for Long-Term Cybersecurity,
from the Berkeley Drive Project, and from Google.




\bibliographystyle{IEEEtranS}
\bibliography{ref.bib}
%



\end{document}